\documentclass{article}

\usepackage{graphicx} 
\usepackage{subfigure} 

\usepackage{natbib}

\usepackage{algorithm}
\usepackage{algorithmic}

\usepackage{hyperref}

\usepackage[accepted]{icml2012}

\usepackage{amsmath}
\usepackage{amsthm}
\usepackage{amssymb}
\usepackage{bm}
\usepackage{multirow}
\usepackage[capitalise]{cleveref}
\newcounter{mnote}

\newcommand{\ie}{i.e.\ }

\newcommand{\chapterabbrv}{}

\newcommand{\flabel}[1]{\label{fig:\chapterabbrv#1}}
\newcommand{\seclabel}[1]{\label{sec:\chapterabbrv#1}}
\newcommand{\tlabel}[1]{\label{tab:\chapterabbrv#1}}
\newcommand{\elabel}[1]{\label{eq:\chapterabbrv#1}}
\newcommand{\alabel}[1]{\label{alg:\chapterabbrv#1}}

\newcommand{\fref}[1]{\cref{fig:#1}}
\newcommand{\sref}[1]{\cref{sec:#1}}
\newcommand{\tref}[1]{\cref{tab:#1}}
\newcommand{\eref}[1]{\cref{eq:#1}}
\newcommand{\aref}[1]{\cref{alg:#1}}

\newcommand*\idx[2][]
{ 
\def\next{#1}%
\ifx\empty\next
  (#2)
\else
  (#1, #2)
\fi
}
\newcommand*\elt[3][]
{ 
\def\next{#1}%
\ifx\empty\next
  #2\idx{#3}
\else
  #1\idx{#2,#3}
\fi
}
\newcommand*\pd[3][]
{ 
\def\next{#1}%
\ifx\empty\next
  \frac{\partial#2}{\partial #3}
\else
  \frac{{\partial^{#1} #2}}{\partial#3^{#1}}
\fi
}

\newcommand{\hfor}{\overrightarrow{h}}
\newcommand{\hback}{\overleftarrow{h}}
\newcommand{\igate}{\alpha}
\newcommand{\fgate}{\beta}
\newcommand{\ogate}{\gamma}
\newcommand{\state}{s}
\newcommand{\hiddenfn}{\mathcal{H}}
\newcommand{\kronecker}[2]{\delta_{#1 #2}}
\newcommand{\len}[1]{|#1|}
\newcommand{\outelt}{k}
\newcommand{\wtmat}[2]{W_{#1 #2}}
\newcommand{\ihwts}{\wtmat{i}{h}}
\newcommand{\hhwts}{\wtmat{h}{h}}
\newcommand{\howts}{\wtmat{h}{o}}
\newcommand{\bias}[1]{b_{#1}}
\newcommand{\hbias}{\bias{h}}
\newcommand{\obias}{\bias{o}}
\newcommand{\seq}[1]{\boldsymbol #1}
\newcommand{\alignment}{\seq{a}}
\newcommand{\bestout}{\best{\outseq}}
\renewcommand{\null}{\varnothing}
\newcommand{\nullseq}{\seq{\null}}
\newcommand{\slice}[3]{#1_{[#2:#3]}}
\newcommand{\loss}{\mathcal{L}}

\newcommand{\invble}{x}
\newcommand{\outvble}{y}
\newcommand{\inseq}{\seq{\invble}}
\newcommand{\outseq}{\seq{\outvble}}
\newcommand{\predinseq}{\hat{\outseq}}

\newcommand{\targseq}{\outseq^*}
\newcommand{\varspace}[1]{\mathcal{\MakeUppercase{#1}}}
\newcommand{\inspace}{\varspace{\invble}}
\newcommand{\outspace}{\varspace{\outvble}}
\newcommand{\nulloutspace}{\bar{\outspace}}
\newcommand{\seqsover}[1]{#1^*}
\newcommand{\outseqspace}{\seqsover{\outspace}}
\newcommand{\nulloutseqspace}{\seqsover{\nulloutspace}}
\newcommand{\range}[3]{#1 \leq #2 \leq #3}
\newcommand{\trange}{\range{1}{t}{T}}
\newcommand{\urange}{\range{0}{u}{U}}

\newcommand{\infn}{f}
\newcommand{\outfn}{g}
\newcommand{\densityfn}{h}

\newcommand{\outprobvar}[3]{\Pr(#1| #2, #3)}
\newcommand{\outprobv}[1]{\outprobvar{#1}{t}{u}}
\newcommand{\outprob}{\outprobv{\outelt}}
\newcommand{\outseqprob}{\Pr(\outseq | \inseq)}
\newcommand{\targseqprob}{\Pr(\targseq | \inseq)}
\newcommand{\outslice}[2]{\slice{\outseq}{#1}{#2}}

\newcommand{\tseq}{\seq{f}}
\newcommand{\tslice}[2]{\slice{\tseq}{#1}{#2}}
\newcommand{\outpref}{\hat{\outseq}}
\newcommand{\outprefs}{pref(\outseq)}
\newcommand{\outlen}{\len{\outseq}}
\newcommand{\outpreflen}{\len{\outpref}}
\newcommand{\outfunc}[2]{\outfn(\outslice{#1}{#2})}
\newcommand{\infunc}[1]{\infn(#1, \inseq)}
\newcommand{\outfuncu}{\outfunc{1}{u}}
\newcommand{\infunct}{\infunc{t}}

\newcommand{\densityfunc}[1]{\densityfn(#1, t, u)}
\newcommand{\density}{\densityfunc{\outelt}}

\newcommand{\fnvals}[2]{#1 \mapsto #2}
\newcommand{\fn}[3]{#1 : \fnvals{#2}{#3}}

\newcommand{\opequals}[1]{\ #1\hspace{-3pt}=\hspace{1pt}}
\newcommand{\plusequals}{\opequals{+}}

\newcommand{\outputvble}{y}
\newcommand{\out}{\seq{\outputvble}}

\newcommand{\aligncollapse}{\mathcal{B}}
\newcommand{\forward}[2]{\elt{\alpha}{#1,#2}}
\newcommand{\backward}[2]{\elt{\beta}{#1,#2}}

\newcommand{\best}[1]{#1^*}

\newcommand{\capt}[2]{\caption[#1.]{\textbf{#1.}#2}}

\newcommand{\figt}[5]
{
\begin{figure}[t]
\begin{center}
\includegraphics[width=#3\columnwidth]{figures/#1}
\end{center}
\capt{#4}{#5}
\flabel{#2}
\end{figure}
}
\newcommand{\figstar}[5]
{
\begin{figure*}
\begin{center}
\includegraphics[width=#3\textwidth]{figures/#1}
\end{center}
\capt{#4}{#5}
\flabel{#2}
\end{figure*}
}

\icmltitlerunning{Sequence Transduction with Recurrent Neural Networks}

\begin{document} 

\twocolumn[
\icmltitle{Sequence Transduction with Recurrent Neural Networks}

\icmlauthor{Alex Graves}{graves@cs.toronto.edu}
\icmladdress{Department of Computer Science, University of Toronto, Canada}

\icmlkeywords{recurrent neural networks, sequence learning, transduction, machine learning, ICML}

\vskip 0.3in
]

\begin{abstract}
Many machine learning tasks can be expressed as the transformation---or \emph{transduction}---of input sequences into output sequences: speech recognition, machine translation, protein secondary structure prediction and text-to-speech to name but a few. One of the key challenges in sequence transduction is learning to represent both the input and output sequences in a way that is invariant to sequential distortions such as shrinking, stretching and translating. Recurrent neural networks (RNNs) are a powerful sequence learning architecture that has proven capable of learning such representations. However RNNs traditionally require a pre-defined alignment between the input and output sequences to perform transduction. This is a severe limitation since \emph{finding} the alignment is the most difficult aspect of many sequence transduction problems. Indeed, even determining the length of the output sequence is often challenging. This paper introduces an end-to-end, probabilistic sequence transduction system, based entirely on RNNs, that is in principle able to transform any input sequence into any finite, discrete output sequence. Experimental results for phoneme recognition are provided on the TIMIT speech corpus.
\end{abstract}

\section{Introduction}
\seclabel{intro}
The ability to transform and manipulate sequences is a crucial part of human intelligence: everything we know about the world reaches us in the form of sensory sequences, and everything we do to interact with the world requires sequences of actions and thoughts.
The creation of automatic sequence transducers therefore seems an important step towards artificial intelligence.
A major problem faced by such systems is how to represent sequential information in a way that is invariant, or at least robust, to sequential distortions.
Moreover this robustness should apply to both the input and output sequences.

For example, transforming audio signals into sequences of words requires the ability to identify speech sounds (such as phonemes or syllables) despite the apparent distortions created by different voices, variable speaking rates, background noise etc.
If a language model is used to inject prior knowledge about the output sequences, it must also be robust to missing words, mispronunciations, non-lexical utterances etc.

Recurrent neural networks (RNNs) are a promising architecture for general-purpose sequence transduction.
The combination of a high-dimensional multivariate internal state and nonlinear state-to-state dynamics offers more expressive power than conventional sequential algorithms such as hidden Markov models.
In particular RNNs are better at storing and accessing information over long periods of time.
While the early years of RNNs were dogged by difficulties in learning~\cite{hochreiter01book}, recent results have shown that they are now capable of delivering state-of-the-art results in real-world tasks such as handwriting recognition~\cite{graves08online,graves09offline}, text generation~\cite{sutskever11rnn} and language modelling~\cite{mikolov10language}.
Furthermore, these results demonstrate the use of long-range memory to perform such actions as closing parentheses after many intervening characters~\cite{sutskever11rnn}, or using delayed strokes to identify handwritten characters from pen trajectories~\cite{graves08online}.

%
%
%
%
%
%

%

However RNNs are usually restricted to problems where the alignment between the input and output sequence is known in advance.
For example, RNNs may be used to classify every frame in a speech signal, or every amino acid in a protein chain.
If the network outputs are probabilistic this leads to a distribution over output sequences of the same length as the input sequence.
But for a general-purpose sequence transducer, where the output length is unknown in advance, we would prefer a distribution over sequences of \emph{all} lengths.
Furthermore, since we do not how the inputs and outputs should be aligned, this distribution would ideally cover all possible alignments.

Connectionist Temporal Classification (CTC) is an RNN output layer that defines a distribution over all alignments with all output sequences not longer than the input sequence~\cite{graves06icml}.
However, as well as precluding tasks, such as text-to-speech, where the output sequence is longer than the input sequence, CTC does not model the interdependencies between the outputs.
The transducer described in this paper extends CTC by defining a distribution over output sequences of all lengths, and by jointly modelling both input-output and output-output dependencies.

As a discriminative sequential model the transducer has similarities with `chain-graph' conditional random fields (CRFs)~\cite{lafferty01crf}.
However the transducer's construction from RNNs, with their ability to extract features from raw data and their potentially unbounded range of dependency, is in marked contrast with the pairwise output potentials and hand-crafted input features typically used for CRFs.
Closer in spirit is the \emph{Graph Transformer Network}~\cite{bottou97transformer} paradigm, in which  differentiable modules (often neural networks) can be globally trained to perform consecutive graph transformations such as detection, segmentation and recognition.

\sref{model} defines the RNN transducer, showing how it can be trained and applied to test data, \sref{experiments} presents experimental results on the TIMIT speech corpus and concluding remarks and directions for future work are given in \sref{conclusion}.

\section{Recurrent Neural Network Transducer}
\seclabel{model}

Let $\inseq = (x_1, x_2,\ldots, x_T)$ be a length $T$ \emph{input sequence} of arbitrary length belonging to the set $\inspace^*$ of all sequences over some \emph{input space} $\inspace$.
Let $\outseq = (y_1, y_2,\ldots, y_U)$ be a length $U$ \emph{output sequence} belonging to the set $\outseqspace$ of all sequences over some \emph{output space} $\outspace$.
Both the inputs vectors $x_t$ and the output vectors $y_u$ are represented by fixed-length real-valued vectors; for example if the task is phonetic speech recognition, each $x_t$ would typically be a vector of MFC coefficients and each $y_t$ would be a one-hot vector encoding a particular phoneme.
In this paper we will assume that the output space is discrete; however the method can be readily extended to continuous output spaces, provided a tractable, differentiable model can be found for $\outspace$.

Define the \emph{extended output space} $\nulloutspace$ as $\outspace \cup \null$, where $\null$ denotes the \emph{null} output.
The intuitive meaning of $\null$ is `output nothing'; the sequence $(y_1, \null, \null, y_2,\null, y_3) \in \nulloutseqspace$ is therefore equivalent to $(y_1, y_2, y_3) \in \outseqspace$.
We refer to the elements $\alignment \in \nulloutseqspace$ as \emph{alignments}, since the location of the null symbols determines an alignment between the input and output sequences.
Given $\inseq$, the RNN transducer defines a conditional distribution $\Pr(\alignment \in \nulloutseqspace|\inseq)$.
This distribution is then collapsed onto the following distribution over $\outseqspace$
\begin{equation}
\Pr(\outseq \in \outseqspace|\inseq) = \sum_{\alignment \in \aligncollapse^{-1}(\outseq)}{\Pr(\alignment|\inseq)}
\end{equation}
where $\fn{\aligncollapse}{\nulloutseqspace}{\outseqspace}$ is a function that removes the null symbols from the alignments in $\nulloutseqspace$.

Two recurrent neural networks are used to determine $\Pr(\alignment \in \nulloutseqspace|\inseq)$.
One network, referred to as the \emph{transcription network} $\mathcal{F}$, scans the input sequence $\inseq$ and outputs the sequence $\seq{f} = (f_1,\ldots,f_T)$ of transcription vectors\footnote{For simplicity we assume the transcription sequence to be the same length as the input sequence; however this may not be true, for example if the transcription network uses a pooling architecture~\cite{lecun98convolution} to reduce the sequence length.}.
The other network, referred to as the \emph{prediction network} $\mathcal{G}$, scans the output sequence $\outseq$ and outputs the prediction vector sequence $\seq{g} = (g_0, g_1\ldots,g_U)$.

\subsection{Prediction Network}
The prediction network $\mathcal{G}$ is a recurrent neural network consisting of an input layer, an output layer and a single hidden layer.
The length $U+1$ input sequence $\predinseq = (\null, y_1,\ldots,y_U)$ to $\mathcal{G}$ output sequence $\outseq$ with $\null$ prepended.
The inputs are encoded as one-hot vectors; that is, if $\outspace$ consists of $K$ labels and $y_{u} = k$, then $\predinseq_u$ is a length $K$ vector whose elements are all zero except the $k^{th}$, which is one.
$\null$ is encoded as a length $K$ vector of zeros.
The input layer is therefore size $K$.
The output layer is size $K+1$ (one unit for each element of $\nulloutspace$) and hence the prediction vectors $g_u$ are also size $K+1$.

Given $\predinseq$, $\mathcal{G}$ computes the hidden vector sequence $(h_0,\ldots,h_U)$ and the prediction sequence $(g_0,\ldots,g_U)$ by iterating the following equations from $u=0$ to $U$:
\begin{align}
\elabel{rnn_hidden}
h_u &= \hiddenfn\left(\ihwts \predinseq_u + \hhwts h_{u-1} + \hbias \right)\\
g_u &= \howts h_u + \obias
\end{align} 
where $\ihwts$ is the input-hidden weight matrix, $\hhwts$ is the hidden-hidden weight matrix, $\howts$ is the hidden-output weight matrix, $\hbias$ and $\obias$ are bias terms, and $\hiddenfn$ is the hidden layer function.
In traditional RNNs $\hiddenfn$ is an elementwise application of the $tanh$ or \emph{logistic sigmoid} $\sigma(x) = 1/(1 + \exp(-x))$ functions.
However we have found that the Long Short-Term Memory (LSTM) architecture~\cite{hochreiter97lstm,gers01thesis} is better at finding and exploiting long range contextual information.
For the version of LSTM used in this paper $\hiddenfn$ is implemented by the following composite function:
\begin{align}
\elabel{lstm1}
\igate_n &= \sigma\left(\wtmat{i}{\igate} i_n + \wtmat{h}{\igate} h_{n-1} + \wtmat{s}{\igate} s_{n-1} \right)\\
\elabel{lstm2}
\fgate_n &= \sigma\left(\wtmat{i}{\fgate} i_n + \wtmat{h}{\fgate} h_{n-1} + \wtmat{s}{\fgate} s_{n-1} \right)\\
\elabel{lstm3}
\state_n &= \fgate_n \state_{n-1} + \igate_n \tanh \left(\wtmat{i}{\state} i_n + \wtmat{h}{\state}\right)\\
\elabel{lstm4}
\ogate_n &= \sigma\left(\wtmat{i}{\ogate} i_n + \wtmat{h}{\ogate} h_{n-1} + \wtmat{s}{\ogate} s_{n} \right)\\
\elabel{lstm5}
h_n &= \ogate_n \tanh(\state_n)
\end{align}
where $\igate$, $\fgate$, $\ogate$ and $\state$ are respectively the \emph{input gate}, \emph{forget gate}, \emph{output gate} and \emph{state} vectors, all of which are the same size as the hidden vector $h$.
The weight matrix subscripts have the obvious meaning, for example $\wtmat{h}{\igate}$ is the hidden-input gate matrix, $\wtmat{i}{\ogate}$ is the input-output gate matrix etc.
The weight matrices from the state to gate vectors are diagonal, so element $m$ in each gate vector only receives input from element $m$ of the state vector.
The bias terms (which are added to $\igate$, $\fgate$, $\state$ and $\ogate$) have been omitted for clarity. 

The prediction network attempts to model each element of $\outseq$ given the previous ones; it is therefore similar to a standard next-step-prediction RNN, only with the added option of making `null' predictions.

\subsection{Transcription Network}

The transcription network $\mathcal{F}$ is a \emph{bidirectional RNN}~\cite{schuster97brnn} that scans the input sequence $\inseq$ forwards and backwards with two separate hidden layers, both of which feed forward to a single output layer.
Bidirectional RNNs are preferred because each output vector depends on the whole input sequence (rather than on the previous inputs only, as is the case with normal RNNs); however we have not tested to what extent this impacts performance.

Given a length $T$ input sequence $(x_1\ldots x_T)$, a bidirectional RNN computes the \emph{forward} hidden sequence $(\hfor_1,\ldots,\hfor_T)$, the \emph{backward} hidden sequence $(\hback_1,\ldots,\hback_T)$, and the transcription sequence $(f_1,\ldots,f_T)$ by first iterating the backward layer from $t=T$ to $1$:
\begin{align}
\hback_t = \hiddenfn\left(\wtmat{i}{\hback} i_t + \wtmat{\hback}{\hback} \hback_{t+1} + \bias{\hback} \right)
\end{align}
then iterating the forward and output layers from $t=1$ to $T$:
\begin{align}
\hfor_t &= \hiddenfn\left(\wtmat{i}{\hfor} i_t + \wtmat{\hfor}{\hfor} \hfor_{t-1} + \bias{\hfor} \right)\\
o_t &= \wtmat{\hfor}{o} \hfor_t + \wtmat{\hback}{o} \hback_t + \obias
\end{align}
For a bidirectional LSTM network~\cite{graves05blstm}, $\hiddenfn$ is implemented by \eref{lstm1,eq:lstm2,eq:lstm3,eq:lstm4,eq:lstm5}.
For a task with $K$ output labels, the output layer of the transcription network is size $K+1$, just like the prediction network, and hence the transcription vectors $f_t$ are also size $K+1$. 

The transcription network is similar to a Connectionist Temporal Classification RNN, which also uses a null output to define a distribution over input-output alignments.

\subsection{Output Distribution}
Given the transcription vector $f_t$, where $1 \leq t \leq T$, the prediction vector $g_u$, where $0 \leq u \leq U$, and label $k \in \nulloutspace$, define the \emph{output density function}
\begin{equation}
	\density = \exp\left(f_t^k + g_u^k\right)
\end{equation}
where superscript $k$ denotes the $k^{th}$ element of the vectors.
The density can be normalised to yield the conditional \emph{output distribution}:
\begin{equation}
\elabel{softmax}
\Pr(k\in\nulloutspace|t,u) = \frac{\density}{\sum_{\outelt' \in \nulloutspace}{h(k',t,u)}}
\end{equation}
To simplify notation, define 
\begin{align}
y(t, u) &\equiv \Pr(y_{u+1}|t, u)\\
\null(t, u) &\equiv \Pr(\null|t, u)
\end{align}
$\outprob$ is used to determine the transition probabilities in the lattice shown in \fref{transducer}.
The set of possible paths from the bottom left to the terminal node in the top right corresponds to the complete set of alignments between $\inseq$ and $\outseq$, \ie to the set $\nulloutseqspace \cap \aligncollapse^{-1}(\outseq)$.
Therefore all possible input-output alignments are assigned a probability, the sum of which is the total probability $\Pr(\outseq|\inseq)$ of the output sequence given the input sequence.
Since a similar lattice could be drawn for \emph{any} finite $\out \in \outseqspace$, $\outprob$ defines a distribution over all possible output sequences, given a single input sequence.

\figt{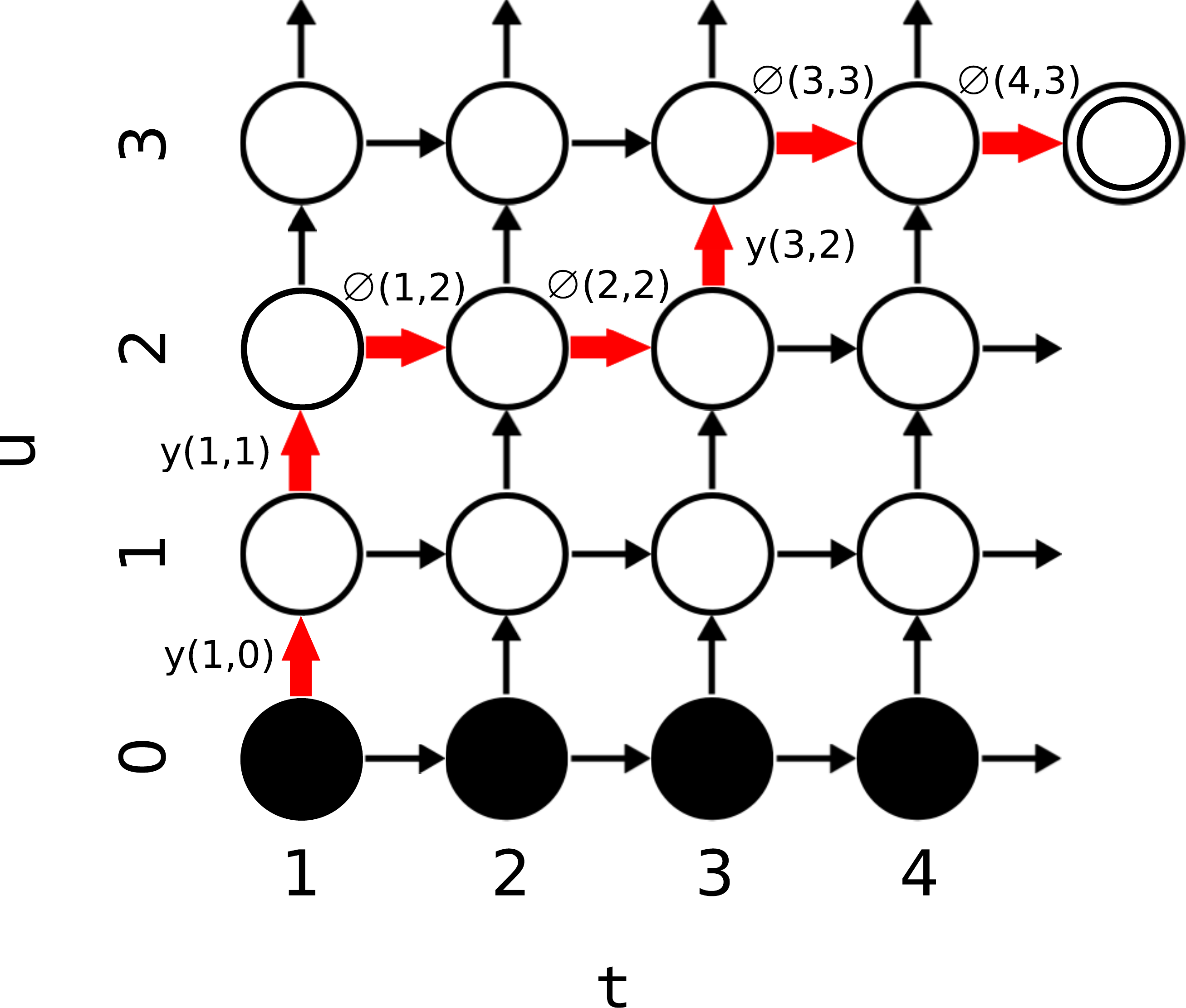}{transducer}{0.9}{Output probability lattice defined by $\outprob$}{ The node at $t, u$ represents the probability of having output the first $u$ elements of the output sequence by point $t$ in the transcription sequence. The horizontal arrow leaving node $t, u$ represents the probability $\null(t, u)$ of outputting nothing at $(t, u)$; the vertical arrow represents the probability $y(t, u)$ of outputting the element $u+1$ of $\outseq$. 
The black nodes at the bottom represent the null state before any outputs have been emitted.
The paths starting at the bottom left and reaching the terminal node in the top right (one of which is shown in red) correspond to the possible alignments between the input and output sequences.
Each alignment starts with probability 1, and its final probability is the product of the transition probabilities of the arrows they pass through (shown for the red path).}

A naive calculation of $\Pr(\outseq|\inseq)$ from the lattice would be intractable; however an efficient forward-backward algorithm is described below.

\subsection{Forward-Backward Algorithm}
\seclabel{forward_backward}
Define the \emph{forward variable} $\forward{t}{u}$ as the probability of outputting $\outslice{1}{u}$ during $\tslice{1}{t}$.
The forward variables for all $\trange$ and $\urange$ can be calculated recursively using
\begin{align}
\elabel{forward}
\forward{t}{u} &= \forward{t-1}{u} \varnothing (t-1, u)\nonumber\\ &+ \forward{t}{u-1} y(t, u-1)
\end{align}
with initial condition $\forward{1}{0} = 1$.
The total output sequence probability is equal to the forward variable at the terminal node:
\begin{equation}
\outseqprob = \forward{T}{U} \varnothing(T, U)
\end{equation}
Define the \emph{backward variable} $\backward{t}{u}$ as the probability of outputting $\outslice{u+1}{U}$ during $\tslice{t}{T}$.
Then
\begin{align}
\elabel{backward}
\backward{t}{u} = \backward{t+1}{u} \varnothing(t, u) + \backward{t}{u+1} y(t, u)
\end{align}
with initial condition $\backward{T}{U} = \varnothing(T, U)$.
From the definition of the forward and backward variables it follows that their product $\forward{t}{u} \backward{t}{u}$ at any point $(t, u)$ in the output lattice is equal to the probability of emitting the complete output sequence \emph{if $y_u$ is emitted during transcription step $t$}.
\fref{forward_backward} shows a plot of the forward variables, the backward variables and their product for a speech recognition task.

\figstar{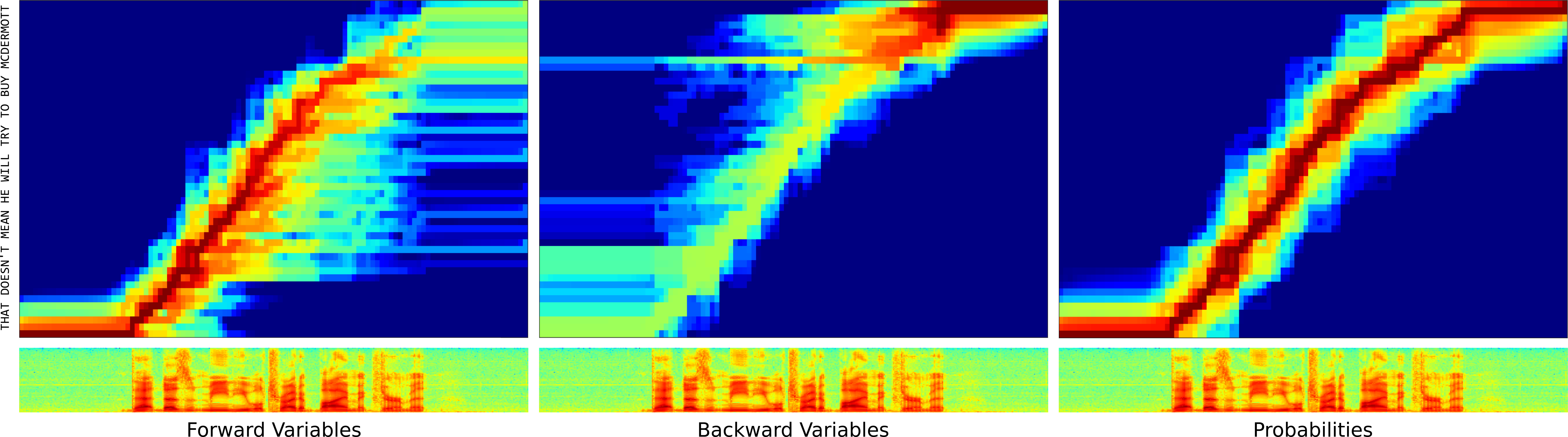}
{forward_backward}{1}{Forward-backward variables during a speech recognition task}{ The image at the bottom is the input sequence: a spectrogram of an utterance. 
The three heat maps above that show the logarithms of the 
forward variables (top) backward variables (middle) and their product (bottom) 
across the output lattice. The text to the left is the target sequence. 
}

\subsection{Training}
Given an input sequence $\inseq$ and a \emph{target sequence} $\targseq$, the natural way to train the model is to minimise the log-loss $\loss = -\ln \targseqprob$ of the target sequence.
We do this by calculating the gradient of $\loss$ with respect to the network weights parameters and performing gradient descent.
Analysing the diffusion of probability through the output lattice shows that $\targseqprob$ is equal to the sum of $\forward{t}{u} \backward{t}{u}$ over any top-left to bottom-right diagonal through the nodes.
That is, $\forall\ n: \range{1}{n}{U+T}$
\begin{equation}
\elabel{seq_prob}
\targseqprob = \hspace{-5pt}\sum_{(t, u) : t+u=n}{\hspace{-10pt}\forward{t}{u}\backward{t}{u}}
\end{equation}
From \eref{forward,eq:backward,eq:seq_prob} and the definition of $\loss$ it follows that
\begin{equation}
\elabel{loss_deriv}
\pd{\loss}{\outprob} = -\frac{\forward{t}{u}}{\targseqprob}
\begin{cases}
\backward{t}{u+1} \text{ if $\outelt = y_{u+1}$}\\
\backward{t+1}{u} \text{ if $\outelt = \null$}\\ 
0 \text{ otherwise}
\end{cases}
\end{equation}
And therefore
\begin{align}
\elabel{infunc_deriv}
\pd{\loss}{f_t^k} &= \sum_{u=0}^U{\sum_{k'\in\nulloutspace}{\pd{\loss}{\outprobv{k'}}\pd{\outprobv{k'}}{f_t^k}}}\\
\pd{\loss}{g_u^k} &= \sum_{t=1}^T{\sum_{k'\in\nulloutspace}{\pd{\loss}{\outprobv{k'}}\pd{\outprobv{k'}}{g_u^k}}}
\end{align}
where, from \eref{softmax}
\begin{align*}
\pd{\outprobv{k'}}{f_t^{k}}\hspace{-1.5pt}=\hspace{-1.5pt}\pd{\outprobv{k'}}{g_u^{k}}\hspace{-1.5pt}=\hspace{-1.5pt}\outprobv{k'}\left[\kronecker{k}{k'}\hspace{-1.5pt}-\hspace{-1.5pt}\outprob\right]
\end{align*}
The gradient with respect to the network weights can then be calculated by applying \emph{Backpropagation Through Time}~\cite{williams95bptt} to each network independently.

A separate softmax could be calculated for every $\outprob$ required by the forward-backward algorithm.
However this is computationally expensive due to the high cost of the exponential function.
Recalling that $\exp(a + b) = \exp(a) \exp(b)$, we can instead precompute all the $\exp\left(\infunct\right)$ and $\exp(\outfuncu)$ terms and use their products to determine $\outprob$.
This reduces the number of exponential evaluations from $O(T U)$ to $O(T + U)$ for each length $T$ transcription sequence and length $U$ target sequence used for training.
\subsection{Testing}
When the transducer is evaluated on test data, we seek the mode of the output sequence distribution induced by the input sequence.
Unfortunately, finding the mode is much harder than determining the probability of a single sequence. 
The complication is that the prediction function $\outfuncu$ (and hence the output distribution $\outprob$) may depend on \emph{all} previous outputs emitted by the model.
The method employed in this paper is a fixed-width beam search through the tree of output sequences.
The advantage of beam search is that it scales to arbitrarily long sequences, and allows computational cost to be traded off against search accuracy.

Let $\Pr(\outseq)$ be the approximate probability of emitting some output sequence $\outseq$ found by the search so far.
Let $\Pr(k|\outseq, t)$ be the probability of extending $\outseq$ by $k \in \nulloutspace$ during transcription step $t$.
Let $\outprefs$ be the set of proper prefixes of $\outseq$ (including the null sequence $\nullseq$), and for some $\outpref \in \outprefs$, let $\Pr(\outseq|\outpref, t) = \prod_{u=\outpreflen + 1}^{\outlen}{\Pr(y_u|\slice{\outseq}{0}{u-1}, t)}$.
Pseudocode for a width $W$ beam search for the output sequence with highest length-normalised probability given some length $T$ transcription sequence is given in \aref{beam_search}.

\begin{algorithm}[h]
\caption{Output Sequence Beam Search}
\alabel{beam_search}
\begin{algorithmic}
\STATE \textbf{Initalise:} $B = \{\nullseq\}$; $\Pr(\nullseq) = 1$
\FOR{$t=1$ \textbf{to} $T$}
\STATE $A = B$
\STATE $B = \{\}$
\FOR{$\outseq$ \textbf{in} $A$}
\STATE $\Pr(\outseq) \plusequals \sum_{\outpref \in \outprefs \cap A}{\Pr(\outpref)\Pr(\outseq|\outpref, t)}$
\ENDFOR
\WHILE{$B$ contains less than $W$ elements more\\\ \ \ \ \ \ \ \ \ probable than the most probable in $A$}
\STATE $\bestout =$ most probable in $A$
\STATE Remove $\bestout$ from $A$
\STATE $\Pr(\bestout) = \Pr(\bestout)\Pr(\null|\outseq, t)$
\STATE Add $\bestout$ to $B$
\FOR{$\outelt \in \outspace$}
	\STATE $\Pr(\bestout + k) = \Pr(\bestout) \Pr(k|\bestout, t)$
	\STATE Add $\bestout + k$ to $A$
\ENDFOR
\ENDWHILE
\STATE Remove all but the $W$ most probable from $B$
\ENDFOR
\STATE \textbf{Return:} $\outseq$ with highest $\log \Pr(\outseq) / |\outseq|$ in $B$ 
\end{algorithmic}
\end{algorithm}

The algorithm can be trivially extended to an $N$ best search ($N \leq W$) by returning a sorted list of the $N$ best elements in $B$ instead of the single best element. 
The length normalisation in the final line appears to be important for good performance, as otherwise shorter output sequences are excessively favoured over longer ones; similar techniques are employed for hidden Markov models in speech and handwriting recognition~\cite{bertolami06rejection}.

Observing from \eref{rnn_hidden} that the prediction network outputs are independent of previous hidden vectors given the current one, we can iteratively compute the prediction vectors for each output sequence $\outseq + \outelt$ considered during the beam search by storing the hidden vectors for all $\outseq$, and running \eref{rnn_hidden} for one step with $k$ as input.
The prediction vectors can then be combined with the transcription vectors to compute the probabilities.
This procedure greatly accelerates the beam search, at the cost of increased memory use.
Note that for LSTM networks both the hidden vectors $h$ and the state vectors $s$ should be stored.

\section{Experimental Results}
\seclabel{experiments}

To evaluate the potential of the RNN transducer we applied it to the task of phoneme recognition on the TIMIT speech corpus~\cite{timit}.
We also compared its performance to that of a standalone next-step prediction RNN and a standalone Connectionist Temporal Classification (CTC) RNN, to gain insight into the interaction between the two sources of information.

\subsection{Task and Data}

The core training and test sets of TIMIT (which we used for our experiments) contain respectively 3696 and 192 phonetically transcribed utterances.
We defined a validation set by randomly selecting 184 sequences from the training set; this put us at a slight disadvantage compared to many TIMIT evaluations, where the validation set is drawn from the non-core test set, and all 3696 sequences are used for training. 
The reduced set of 39 phoneme targets~\cite{lee89timit39} was used during both training and testing.

Standard speech preprocessing was applied to transform the audio files into feature sequences. 26 channel mel-frequency filter bank and a pre-emphasis coefficient of 0.97 were used to compute 12 mel-frequency cepstral coefficients plus an energy coefficient on 25ms Hamming windows at 10ms intervals. 
Delta coefficients were added to create input sequences of length 26 vectors, and all coefficient were normalised to have mean zero and standard deviation one over the training set.

The standard performance measure for TIMIT is the phoneme error rate on the test set: that is, the summed edit distance between the output sequences and the target sequences, divided by the total length of the target sequences.
Phoneme error rate, which is customarily presented as a percentage, is recorded for both the transcription network and the transducer.
The error recorded for the prediction network is the misclassification rate of the next phoneme given the previous ones.

We also record the log-loss on the test set.
To put this quantity in more accessible terms we convert it into the average number of bits per phoneme target.

\subsection{Network Parameters}

The prediction network consisted of a size 128 LSTM hidden layer, 39 input units and 40 output units.
The transcription network consisted of two size 128 LSTM hidden layers, 26 inputs and 40 outputs. 
This gave a total of 261,328 weights in the RNN transducer.
The standalone prediction and CTC networks (which were structurally identical to their counterparts in the transducer, except that the prediction network had one fewer output unit) had 91,431 and 169,768 weights respectively.
All networks were trained with online steepest descent (weight updates after every sequence) using a learning rate of $10^{-4}$ and a momentum of 0.9.
Gaussian weight noise~\cite{chuen96noise} with a standard deviation of $0.075$ was injected during training to reduce overfitting.
The prediction and transduction networks were stopped at the point of lowest log-loss on the validation set; the CTC network was stopped at the point of lowest phoneme error rate on the validation set.
All network were initialised with uniformly distributed random weights in the range [-0.1,0.1].
For the CTC network, prefix search decoding~\cite{graves06icml} was used to transcribe the test set, with a probability threshold of 0.995.
For the transduction network, the beam search algorithm described in \aref{beam_search} was used with a beam width of 4000.

\subsection{Results}

The results are presented in \tref{timit}.
The phoneme error rate of the transducer is among the lowest recorded on TIMIT (the current benchmark is 20.5\%~\cite{dahl10timit}).
As far as we are aware, it is the best result with a recurrent neural network.

Nonetheless the advantage of the transducer over the CTC network on its own is relatively slight.
This may be because the TIMIT transcriptions are too small a training set for the prediction network: around 150K labels, as opposed to the millions of words typically used to train language models.
This is supported by the poor performance of the standalone prediction network: it misclassifies almost three quarters of the targets, and its per-phoneme loss is not much better than the entropy of the phoneme distribution (4.6 bits).
We would therefore hope for a greater improvement on a larger dataset.
Alternatively the prediction network could be pretrained on a large `target-only' dataset, then jointly retrained on the smaller dataset as part of the transducer.
The analogous procedure in HMM speech recognisers is to combine language models extracted from large text corpora with acoustic models trained on smaller speech corpora.

\begin{table}
\centering
\capt{Phoneme Recognition Results on the TIMIT Speech Corpus}{ `Log-loss' is in units of bits per target phoneme. `Epochs' is the number of passes through the training set before convergence.}
\tlabel{timit}
\vskip 0.15in
\begin{center}
\begin{small}
\begin{sc}\begin{tabular}{llll}
\hline
\abovespace\belowspace
Network & Epochs & Log-loss & Error Rate\\
\hline
\abovespace
Prediction & 58 & 4.0 & 72.9\%\\
CTC & 96 & 1.3 & 25.5\%\\
\belowspace
Transducer & 76 & 1.0 & 23.2\%\\
\hline
\end{tabular}
\end{sc}
\end{small}
\end{center}
\vskip -0.1in
\end{table}

\subsection{Analysis}
One advantage of a differentiable system is that the sensitivity of each component to every other component can be easily calculated.
This allows us to analyse the dependency of the output probability lattice on its two sources of information: the input sequence and the previous outputs.
\fref{jacobian} visualises these relationships for an RNN transducer applied to `end-to-end' speech recognition, where raw spectrogram images are directly transcribed with character sequences with no intermediate conversion into phonemes.

%
%

\begin{figure*}[t]
\begin{center}
\includegraphics[width=\textwidth]{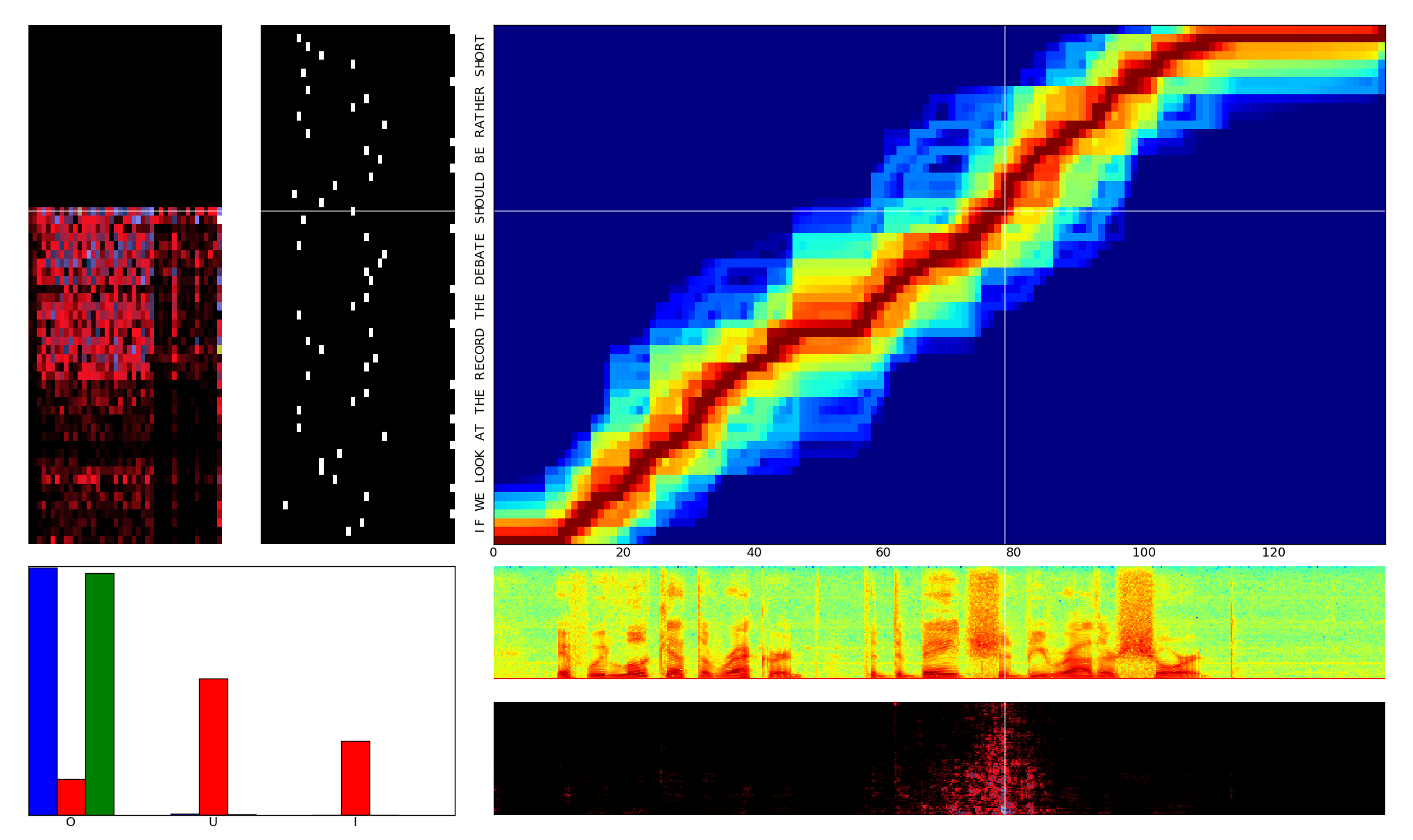}
\end{center}
\capt{Visualisation of the transducer applied to end-to-end speech recognition}{ 
As in \fref{forward_backward}, the heat map in the top right shows the log-probability of the target sequence passing through each point in the output lattice. 
The image immediately below that shows the input sequence (a speech spectrogram), and the image immediately to the left shows the inputs to the prediction network (a series of one-hot binary vectors encoding the target characters).
Note the learned `time warping' between the two sequences.
Also note the blue `tendrils', corresponding to low probability alignments, and the short vertical segments, corresponding to common character sequences (such as `TH' and `HER') emitted during a single input step.
\\
\\
The bar graphs in the bottom left indicate the labels most strongly predicted by the output distribution (blue), the transcription function (red) and the prediction function (green) at the point in the output lattice indicated by the crosshair. In this case the transcription network simultaneously predicts the letters `O', `U' and `L', presumably because these correspond to the vowel sound in `SHOULD'; the prediction network strongly predicts `O'; and the output distribution sums the two to give highest probability to `O'. 
\\
\\
The heat map below the input sequence shows the sensitivity of the probability at the crosshair to the pixels in the input sequence; the heat map to the left of the prediction inputs shows the sensitivity of the same point to the previous outputs. The maps suggest that both networks are sensitive to long range dependencies, with visible effects extending across the length of input and output sequences. Note the dark horizontal bands in the prediction heat map; these correspond to a lowered sensitivity to spaces between words. Similarly the transcription network is more sensitive to parts of the spectrogram with higher energy. The sensitivity of the transcription network extends in both directions because it is bidirectional, unlike the prediction network.}
\flabel{jacobian}
\end{figure*}

\section{Conclusions and Future Work}
\seclabel{conclusion}
We have introduced a generic sequence transducer composed of two recurrent neural networks and demonstrated its ability to integrate acoustic and linguistic information during a speech recognition task.

We are currently training the transducer on large-scale speech and handwriting recognition databases.
Some of the illustrations in this paper are drawn from an ongoing experiment in end-to-end speech recognition.

In the future we would like to look at a wider range of sequence transduction problems, particularly those that are difficult to tackle with conventional algorithms such as HMMs.
One example would be text-to-speech, where a small number of discrete input labels are transformed into long, continuous output trajectories.
Another is machine translation, which is particularly challenging due to the complex alignment between the input and output sequences.

\subsection*{Acknowledgements}
Ilya Sutskever, Chris Maddison and Geoffrey Hinton provided helpful discussions and suggestions for this work. Alex Graves is a Junior Fellow of the Canadian Institute for Advanced Research.

\setlength{\bibsep}{0.0pt}
\bibliography{transducer}
\bibliographystyle{icml2012}

\end{document}